\documentclass[runningheads]{llncs}
\usepackage[T1]{fontenc}

\usepackage{times}
\usepackage{latexsym}
\usepackage{amsmath}
\usepackage{amssymb}
\usepackage{xcolor}
\usepackage{colortbl}

\usepackage[T1]{fontenc}

\usepackage[utf8]{inputenc}

\usepackage{microtype}
\usepackage{dsfont}
\newcommand{\mymodel}{\textsc{EmbRAG}}

\usepackage[most]{tcolorbox}
\definecolor{block-gray}{gray}{0.85}
\newtcolorbox{myquote}{colframe=black,boxrule=1pt,
colback=white,grow to right by=-1mm,grow to left by=-1mm,
boxsep=0pt,breakable}

\newtcolorbox{inside_myquote}{boxrule=0pt,
colback=block-gray,grow to right by=3mm,grow to left by=3mm,
top=0pt,bottom=0pt}

\title{Multi-hop Reasoning and Retrieval in Embedding Space: Leveraging Large Language Models with Knowledge Graphs}

\titlerunning{Multi-hop Reasoning and Retrieval-Augmented Generation in Embedding Space}

\author{
Lihui Liu\inst{1}
}

\authorrunning{Lihui Liu.}

\institute{
Wayne State University, Detroit, Michigan, USA \\
\email{\{hw6926\}@wayne.edu} \\
}

\begin{document}
\maketitle
\begin{abstract}

As large language models (LLMs) continue to grow in size, their abilities to tackle complex tasks have significantly improved. However, issues such as hallucination and the lack of up-to-date knowledge largely remain unresolved. Knowledge graphs (KGs), which serve as symbolic representations of real-world knowledge, offer a reliable source for enhancing reasoning. Integrating KG retrieval into LLMs can therefore strengthen their reasoning by providing dependable knowledge. Nevertheless, due to limited understanding of the underlying knowledge graph, LLMs may struggle with queries that have multiple interpretations. Additionally, the incompleteness and noise within knowledge graphs may result in retrieval failures. To address these challenges, we propose an embedding-based retrieval reasoning framework \mymodel. In this approach, the model first generates multiple logical rules grounded in knowledge graphs based on the input query. These rules are then applied to reasoning in the embedding space, guided by the knowledge graph, ensuring more robust and accurate reasoning. A reranker model further interprets these rules and refines the results. Extensive experiments on two benchmark KGQA datasets demonstrate that our approach achieves the new state-of-the-art performance in KG reasoning tasks.

\end{abstract}

\section{Introduction}

As large language models (LLMs) continue to grow in size, they have shown outstanding performance in many different tasks, such as text generation ~\cite{topal2021exploring}, question answering ~\cite{tan2023can}, fact checking ~\cite{tang2024minicheck} and so on. Recently, more powerful models have come which can solve even more complex tasks with higher accuracy. Popular models like GPT-4 ~\cite{achiam2023gpt} and LLaMA 3 ~\cite{touvron2023llama} have set new records in the field, showcasing the high potential of large-scale neural networks in language understanding and generation.

Despite the tremendous progress, one major issue of LLMs is that it lacks up to date knowledge. Since existing LLMs are trained on past datasets, they do not have the access to the most recent information. 
Moreover, LLMs are prone to hallucination when they answer question that they are unfamiliar with ~\cite{touvron2023llama}. In such cases, they could generate incorrect or even nonsensical answers that do not align with factual knowledge. This unpredictability in generating reliable responses is a significant challenge, seriously hindering the applications on many different tasks, particularly for tasks that require up-to-date and precise information.

To address these issues, several methods have been proposed, and one promising method is Retrieval-Augmented Generation (RAG) \cite{ke2024development}. 
The general philosophy of Retrieval-Augmented Generation (RAG) is to utilize external knowledge to enhance the model's ability to generate accurate information and mitigate hallucination problem. 
By integrating retrieval mechanisms, LLMs can access real-time knowledge and reduce hallucinations, thereby improving the quality of the generated output.

Despite the success of RAG, utilizing knowledge graph during the retrieval process is still challenging. 
One key issue is representing the input query as grounded structures in the knowledge graph. Unlike text retrieval, KGs rely on structured data, whereas each free-form natural language  query can be supported by different substructures of the knowledge graph, making it difficult for RAG to retrieve the correct answer with a single fixed graph query.  
Additionally, KGs themselves are often incomplete, and thus direct semantic matching or graph traversal might fail if relevant edges are missing. These issues highlight the need for more flexible retrieval and reasoning approaches on KGs.

To address these issues, in this paper, we present a novel approach that integrates large language models (LLMs) with knowledge graphs (KGs) for reasoning in the embedding space. To combat diverse structures in the knowledge graph, we propose a logic rule generator that generates different logic rules capable of interpreting the input question during the reasoning process. The process begins with a module that generates multiple logic rules grounded in KGs. These logic rules are then refined to enhance their quality before being employed to retrieve relevant information from the KG. To mitigate the incompleteness of the knowledge graph, we search in the embedding space for answers according to likelihood scores, rather than directly traversing the knowledge graph. At each time step, we compare the embedding information with the ground knowledge graph to ensure that the correct answers consistently have the highest likelihood scores. After that, we utilize large language models (LLMs) along with the logic rules to reason and rerank the results. Extensive experiments conducted on various benchmark KGQA datasets demonstrate that our approach achieves new state-of-the-art performance, producing accurate and interpretable reasoning results.

In summary, the main contributions of this paper are: 
\begin{itemize} 
\item \textbf{Algorithm}: We propose \mymodel, which integrates logic rules and retrieval-augmented generation in embedding space to enhance multi-hop question answering and mitigate the hallucination problem of large language models (LLMs). 
\item \textbf{Empirical Evaluations}: We conducted extensive experiments on several real-world datasets. The results of our experiments demonstrate the effectiveness of \mymodel. 
\end{itemize}


\section{Preliminary and Problem Definition}

Knowledge Graphs serve as structured repositories of information, organizing vast amounts of data into a format that highlights relationships between various entities ~\cite{bordes2013translating}. A knowledge graph can be denoted as $\mathcal{G}=(\mathcal{E}, \mathcal{R}, \mathcal{L})$ where $\mathcal{E} = \{e_1, e_2, ..., e_n\}$ is the set of nodes/entities, $\mathcal{R} = \{r_1, r_2, ..., r_m\}$ is the set of relations and $\mathcal{L}$ is the list of triples.
Each piece of knowledge within a KG is represented as a triple:
$(h, r, t)$. 
In this representation, $h$ and $t$ denote specific entities, while $r$ signifies the relationship connecting them. 
A multi-hop question is represented as a sequence of words $Q = (w_1, w_2, \dots, w_{|Q|})$, with a topic entity $v_Q \in \mathcal{V}$. Each multi-hop query can be mapped to a multi-hop path in the knowledge graph. A multi-hop path $p$ consists of a sequence of relations $p = \{r_1, r_2, \dots, r_l\}$, where each $r_i \in \mathcal{R}$ represents the $i$-th relationship in the path, and $l$ denotes the total number of relations traversed. For example, the question "Who is the child of Tom's brother?" can be represented as the path "\textrm{hasBrother}$\to$\textrm{hasChild}".

Answering a multi-hop query typically requires traversing several links within the KG to arrive at a solution. However, real-world knowledge graphs are usually incomplete, often with missing edges. As a result, directly traversing the graph along a path $p$ may fail to find answers ~\cite{zhang2022learning}. 
On the other hand, due to the complexity of knowledge graphs, each multi-hop question can be supported by multiple logical rules. For example, the question "Who is the child of Tom's brother?" can be interpreted through several different logic rules, such as:
$\textrm{hasBrother} \rightarrow \textrm{hasSpouse} \rightarrow \textrm{hasChild}$ or $\textrm{hasBrother} \rightarrow \textrm{isMarriedTo} \rightarrow \textrm{hasChild}$
or $\textrm{hasBrother} \rightarrow \textrm{isMarriedTo} \rightarrow \textrm{hasFather}^{-1}$. 
This demonstrates the flexibility in reasoning rules within the knowledge graph, where the logic rule is defined as follows. 

\noindent\textbf{Logic Rule.} A logic rule can be represented in the conjunctive form of multiple atomic formulas as follows:
$
\forall \{X_i\}_{i=0}^l \; p(X_0, X_l) \leftarrow r_1(X_0, X_1) \land \cdots \land r_l(X_{l-1}, X_l),
$ where \( p() \) is the rule head corresponding to the path of a multi-hop query, and $r_1(X_0, X_1) \land \cdots \land r_l(X_{l-1}, X_l)$ is the rule body where \( r_i \in \mathcal{R} \) is a binary predicate, with \( l \) denoting the length of the rule.
\vspace{-1.5\baselineskip}
\begin{figure*}[hbt!]
	\centering
	\includegraphics[width=0.84\textwidth]{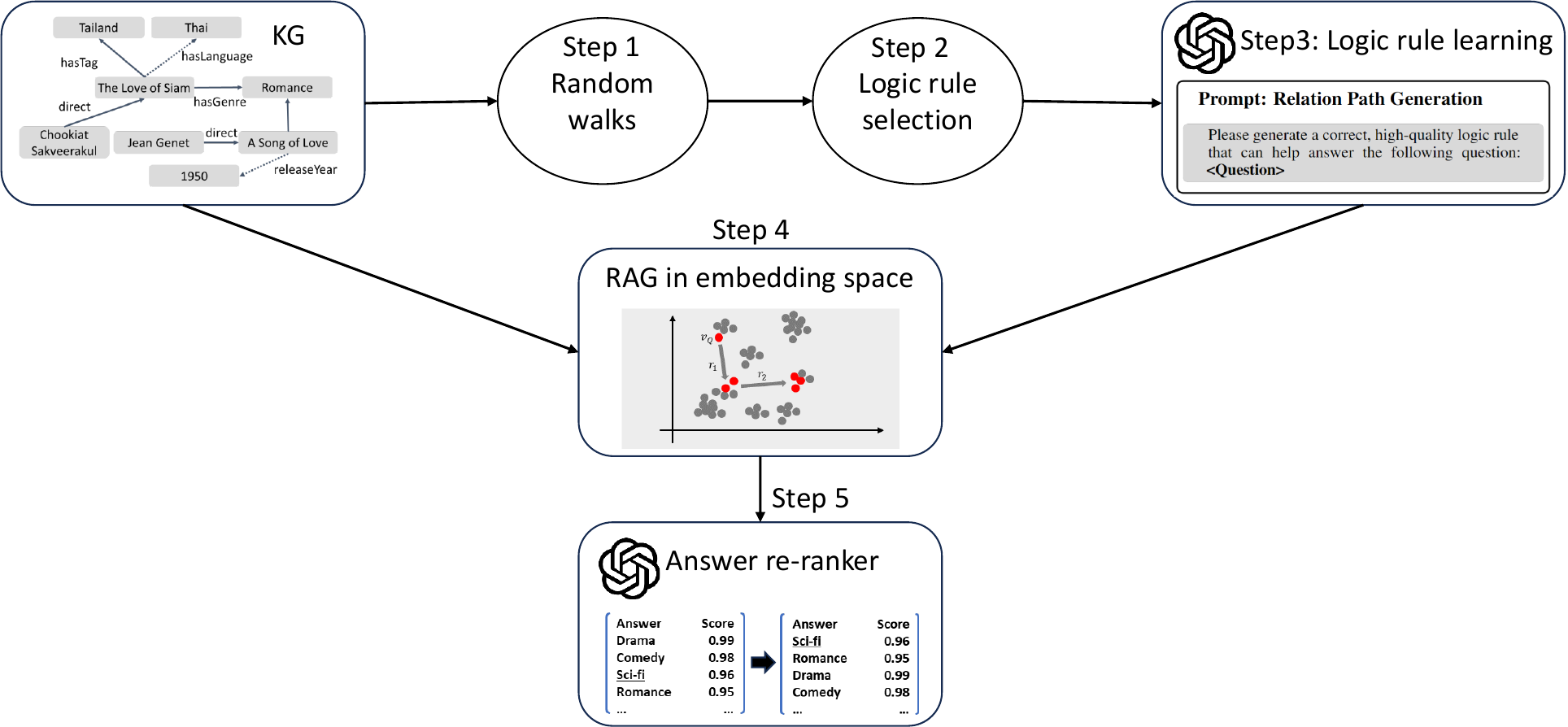}
        
	\caption{Architecture of our proposed
        \label{fig:framework}
 \mymodel\ Model.
}
\vspace{-1.5\baselineskip}
\end{figure*}

Answering multi-hop question aims to find answer in the KG according to various logic rules of the question. It can be treated as an optimization problem that aims to maximize the probability of reasoning the answer from a knowledge graph $G$ with respect to the question $q$ by logic rules $z$: $P(a|q, G) = \sum_{z \in Z} P(a|q, z, G) P(z|q)$,
where $Z$ denotes the set of possible logic rules. 
The latter term $P(z|q)$ is the probability of generating a faithful logic rule $z$ grounded by the knowledge graph $G$, given the question $q$. The former term $P(a|q, z, G)$ is the probability of reasoning an answer $a$ given the question $q$, logic rule $z$, and knowledge graph $G$.

Thus, given a multi-hop query, the goal of this paper is to identify a set of logic rules and utilize them to generate answers by LLMs through retrieval-augmented generation. Formally, the problem definition is

\textbf{Given:} (1) A Knowledge Graph (KG) structured as triples representing relationships between entities, (2) a large language model (LLM), (3) a multi-hop question; 
\textbf{Output:} the relevant answer obtained by effectively navigating multiple relationships in the KG using embedding techniques to enhance reasoning capabilities.

\section{Method}

In this section, we introduce our method, \mymodel, which consists of three key components. First, given a multi-hop question, the Logical Rule Inference Module generates high-quality logic rules grounded in the knowledge graph with respect to the query. Second, the Retrieval-Reasoning Module retrieves valid answers from the knowledge graph by performing reasoning in the embedding space, guided by the generated logic rules. Finally, the Reranking Module selects the most accurate answers based on the retrieved information. The overall framework of \mymodel\ is depicted in Figure~\ref{fig:framework}.

\subsection{Logic Rule Inference}

Given a multi-hop query, many logic rules in the KG might lead us to find the answers. 
For example, the question "Who is the child of Tom's brother?" can be interpreted through several different logic rules, such as:
$\textrm{hasBrother} \rightarrow \textrm{hasSpouse} \rightarrow \textrm{hasChild}$ or $\textrm{hasBrother} \rightarrow \textrm{isMarriedTo} \rightarrow \textrm{hasChild}$
or $\textrm{hasBrother} \rightarrow \textrm{isMarriedTo} \rightarrow \textrm{hasFather}^{-1}$. 
The goal of logic rule inference is to mine rules from KG which can represent the question. Inspired by ~\cite{pra}, given a question and answer set pair, we treat the random walks between the topic entity and the answers as the potential logic rules of the multi-hop question. 

\textbf{Definition I (Random Walk)} 
A random walk $W$ starting from subject entity $e_s \in E$ to object entity $e_o \in E$ in the knowledge graph $\mathcal{G}$ is defined as a sequence of edges 
$$((e_s, r_1, e_1), (e_1, r_2, e_2), \dots, (e_{n-1}, r_n, e_o))$$ 
where $(e_i, r_{i+1}, e_{i+1}) \in \mathcal{G}$ for $i = 0, \dots, n-1$. Each $r_i$ represents a relation between the entities, and the entities $e_1, \dots, e_{n-1}$ are intermediate nodes in the graph. In a random walk, the next edge is chosen randomly from the outgoing edges from the current entity.

After we find all random walks with length up to $L$ between the topic entity and the answers, We use Bayes' Rule to infer the probability of whether the random walk is the correct logic rule. 
First, all the multi-hop questions in the training set can be divided into $m$ clusters. $C_1, C_2, .., C_m$ where $m$ is the number of unique question contexts after masking the topic entity. Each cluster has a corresponding answer superset $\textrm{SUP}_i = \{A_{Q_{j}} | Q_{j} \in C_i\}$. 
Each answer entity $v_i \in A_{Q_{j}}$ has a corresponding KG grounded random walk set $\textrm{RW}(Q_{j}, v_i) = \{p_i | (v_{Q_{j}}, p_i, v_i) \in \mathcal{G}\}$. 
If we assume the occurrence of different random walks in $\textrm{RW}$ as i.i.d. random variables, the probability that this random walk is the correct logic rule can be expressed as $Pr(p_i | C_j, \theta) = \frac{\sum_{Q_j \in C_j} Pr(p_i, Q_j | \theta)}{|\textrm{SUP}_j|}$.
\noindent where $|\textrm{SUP}_j|$ is the number of answer sets in $\textrm{SUP}_j$,
$Q_j \in C_j$ denotes that a specific question $Q_j$ (e.g., ``{\em Who is the child of Tom's brother?}'') belongs to $C_j$ (e.g., ``{\em Who is the child of [NE]'s brother?}'')
and $Pr(p_i, Q_j | \theta)$ is the probability that $p_i \in \textrm{RW}(Q_{j}, v_i)$ for any $v_i \in A_{Q_{j}}$, which is defined as follows, where $\mathds{1}()$ is the indicator function. 
\begin{equation}\label{choose_path}
{\tiny
Pr(p_i, Q_j | \theta) = \frac{\sum_{v_i \in A_{Q_{j}}} |\textrm{RW}(Q_j, v_i)|^{\mathds{1}(p_i \in \textrm{RW}(Q_j, v_i))}}{|A_{Q_{j}}|}
}
\end{equation}
After calculating the probability for all the potential random walks, we select the random walk with the highest probability of each answer as the potentially correct logic rule.

\noindent{\bf Logic Rule Selection.}~
After generating a set of potential logic rules, we use prompt engineering with large language models (LLMs) to select high-quality rules. By leveraging few-shot prompting, we guide LLMs to produce responses that align with the desired format and reasoning process. Previous research has demonstrated that LLMs can learn to respond to prompts and reveal knowledge acquired during pretraining. Inspired by the chain-of-thought prompting technique~\cite{wei2022chain}, we propose a $k$-shot prompting paradigm, where a few examples with reasoning steps are provided to the model to illustrate the selection of high-quality logic rules. This approach enables the model to reason through the options and effectively identify the most appropriate rules. By refining the pool of potential logic rules, we ensure that only the most relevant and accurate ones are retained.

\begin{myquote}

Prompt: Logic Rule Selection

\begin{inside_myquote}\small
Please select correct logic rules that can be used to answer the following multi-hop question: <Question>

Candidate logic rules:

1. <RULE>$r_1$<SEP>$r_2$<SEP>...<SEP>$r_l$</RULE>

2. <RULE>$r_a$<SEP>$r_b$<SEP>...<SEP>$r_m$</RULE>

...

Reason:

...

The correct logic rules are: 
\end{inside_myquote}

\end{myquote}
Where <Question> represents the question $q$. The question, along with the prompt template, is fed into the LLMs to select the logic rules, which are presented in a structured sentence format.
Here, the special tokens \texttt{<RULE>}, \texttt{<SEP>}, and \texttt{</RULE>} indicate the start, separator, and end of the logic rule, respectively.
The selected rules are then treated as the true logic rules.

\subsection{Logic Rule Learning}
In the rule learning process, the goal is to guide the language model to generate high-quality rules based on the question. We utilize instruction tuning to achieve this. By leveraging the instruction-following ability of LLMs~\cite{zeng2023evaluating}, we design a simple instruction template that prompts LLMs to generate logic rules:

\begin{myquote}

\textbf{Prompt: Relation Path Generation}

\begin{inside_myquote}\small
Please generate a correct, high-quality logic rule that can help answer the following question: \textbf{<Question>}
\end{inside_myquote}

\end{myquote}
Where \textrm{<Question>} represents the question $q$. The question, along with the instruction template, is fed into the LLMs to generate the logic rules, which are inferred by the logic rule inference model.
The objective function for instruction tuning can be written as:
\begin{equation*}
\begin{aligned}
\arg \max_{\theta} \frac{1}{|Z^*|} \sum_{z \in Z^*} \log P_{\theta}(z|q) 
= \frac{1}{|Z^*|} \sum_{z \in Z^*} \log \prod_{i=1}^{|z|} P_{\theta}(r_i | r_{<i}, q)
\end{aligned}
\end{equation*}
where $P_{\theta}(z|q)$ represents the likelihood of generating a faithful logic rule $z$, and $P_{\theta}(r_i | r_{<i}, q)$ is the probability of each token in $z$ generated by the LLMs.

\subsection{Answer Retrieval}

When answering multi-hop natural language questions, given a logic rule $z$, the retrieval module aims to extract reasoning paths $p_z$ from the knowledge graph $\mathcal{G}$. The retrieval process is conducted by identifying paths in $\mathcal{G}$ that start from the question entity $e_q$ and follow the logic rule $z$. However, traditional knowledge graph traversal or subgraph matching methods ~\cite{jin2024large} are impractical due to the incompleteness of the knowledge graph.

To address this, we propose a method that finds answers in the embedding space.
Inspired by fuzzy logic ~\cite{kosko1993fuzzy}, we apply fuzzy logic to model this process. A fuzzy logic function can be defined as $\Theta: L \rightarrow [0, 1]$, where $L$ represents an atomic logic formula $r(x, y)$, and the output of this function denotes the probability that the logic formula is true. If the output is restricted to $0$ or $1$, it becomes equivalent to Boolean logic.

Given a logic rule $z = (r_1 \land \dots \land r_l)$ starting from the topic entity $v_Q$, the goal is to predict the logic probability $\Theta: (r_1 \land \dots \land r_l, v) \rightarrow [0, 1]$, which captures the likelihood of following a specific logic sequence from $v_Q$ to any node in $A_Q$. Due to the transition property of logical operations, $\Theta: (r_1 \land \dots \land r_l, v)$ can be expressed as $\Theta(r_1, v_1) \land \Theta(r_2, v_2) \land \dots$. Thus, the final fuzzy logic probability can be computed as:
\begin{equation}\label{path_probability}
\begin{aligned}
\Theta: (r_1, \land..., r_{l}, v) = \Theta(r_1, v_1) \land \cdot \land \Theta(r_l, v_l) 
\end{aligned}
\end{equation}
Specifically, we compute the probability of a target entity $v$ by multiplying the fuzzy probabilities of all intermediate steps traversed by path $P$ in the knowledge graph $\mathcal{G} = (\mathcal{V}, \mathcal{R}, \mathcal{L})$. $\Theta: (r_1, \land..., r_{l}, v) = \prod_{i=1}^{|P|} \Theta(r_i, v_i | P_{1\rightarrow i-1}, v_Q, \mathcal{G})$.
The correct answer is treated as the entity with the highest probability: $o = \textrm{max}_{v_i \in \mathcal{V}} (Pr(v_i | P, v_Q, \mathcal{G}))$.

Let $o(q, e)$ be the embedding model's scoring function, estimating the probability that an entity $e$ is the answer to the query $q$. We aim to use $o()$ to approximate the truth value function $I()$, where $I(q[e])$ evaluates the truth value of a logical formula where $e$ fills the query $q$. Thus, by using logical laws, we can deduce reasonable properties that a query embedding model should possess.

Various methods can be used to model $\Theta$, such as bilinear transformations ~\cite{yang2014embedding}, convolutional networks ~\cite{dettmers2018convolutional}, and others. In our experiments, we use ComplEx ~\cite{trouillon2016complex}. Given $v_h, v_t \in \mathcal{V}$ and $r_i \in \mathcal{R}$, with their corresponding embeddings $\mathbf{e}_h, \mathbf{e}_t, \mathbf{r}_i$, the probability score of $v_t$ being reachable from $v_h$ via $r_i$ is calculated as follows:

\begin{equation}\label{single_probability} \small
Pr(v_t | r_i, v_h, \mathcal{G}) = 
\begin{cases} Re(\langle \textbf{r}_i, \textbf{e}_h, \overline{\textbf{e}}_t \rangle) & (v_h, r_i, v_t) \notin \mathcal{L}, \\ 
1 & (v_h, r_i, v_t) \in \mathcal{L}, 
\end{cases} 
\end{equation}

where $Re()$ represents the real part of the ComplEx ~\cite{trouillon2016complex} output.

\subsection{Answer Re-ranker}

After we retrieve multiple reasoning paths, a straightforward approach would be to use majority voting to determine the answers \cite{minkov2010improving}. However, this method presents several issues, primarily because it can include noisy and irrelevant paths that distort the results \cite{liu2022tlogic}. Thus, we need a more refined strategy to pinpoint the most reliable answers. We propose implementing an effective filtering mechanism that prioritizes accuracy and relevance, ensuring that our final answers are based on solid reasoning paths.

\textbf{Reasoning with Logic Rules.} The goal of the reasoning module is designed to process the question \( q \) and a set of retrieved reasoning paths \( W_z \) to generate the correct answer \( a \). The reasoning process is guided by logic reasoning rules applied to the paths \( W_z \), which are formatted as structured, sequential statements. By leveraging the knowledge encoded in LLMs, the model can filter out noisy or irrelevant paths and focus on those that follow valid logical reasoning patterns.
More specifically, an instruction prompt further guides large language models (LLMs) in reasoning over these paths by emphasizing the alignment of the retrieved logic reasoning rules with the question. This ensures that the generated answers are both relevant and grounded in correct reasoning.

\textbf{Parameter-efficient Instruction Tuning.}~
Direct fine-tuning of the entire model can be both computationally demanding and time-consuming. To address these challenges, we utilize the Low-Rank Adaptation (LoRA) technique \cite{hu2021lora}. LoRA involves freezing the parameters of the pre-trained model while introducing additional trainable parameters that can be expressed as low-rank matrices. This approach efficiently incorporates supplementary information into the language model without requiring extensive computational resources.
Currently, numerous large language models (LLMs) have been released, including popular series like GPT \cite{achiam2023gpt}, T5 \cite{lehman2023we}, Chinchilla \cite{hoffmann2022training}, and LLaMA \cite{touvron2023llama}. However, many proprietary models, such as ChatGPT, are accessible only via APIs, which limits their adaptability for research purposes. To advance our research in generative forecasting for multi-hop question answering, we have chosen open-source LLMs, specifically LLaMA2-7B, as they allow for greater flexibility in adaptation and alignment.

\section{Experiments}

\begin{table*}[hbt!]
\centering
\small
\setlength{\tabcolsep}{1pt} 
\caption{Hits@1 Performance comparison with different baselines on the QA datasets.}
\label{tableRes}
\begin{tabular}{|l|l|c|c|c|c|c|}
\hline
\textbf{Type} & \textbf{Methods} & \textbf{WebQSP} & \textbf{CWQ} & \textbf{MetaQA-1hop} & \textbf{MetaQA-2hop} & \textbf{MetaQA-3hop} \\
\hline
{Retrieval}   & GraftNet          & 66.4         & 36.8          & 97.0          & 94.8            & 77.7           \\
              & PullNet           & {68.1}         & {45.9}          & 97.0          & \underline{99.9 }           & 91.4           \\ \hline
{Embedding}   & KV-Mem            & 46.7         & 18.4          & 96.2          & 82.7            & 48.9           \\
              & EmbedKGQA         & 66.6         & {45.9}          & \textbf{97.5}          & 98.8            & 94.8           \\
              & TransferNet       & {71.4}         & {48.6}          & \textbf{97.5}          & \textbf{100}             & \textbf{100}            \\
              & BiNet             & 67.6         & 47.1          & \textbf{97.5}          & 98.8            & 94.8           \\ \hline
{LLMs}        & ChatGPT           & 66.8         & 39.9          & 66.1          & 45.4            & 41.6           \\
              & ChatGPT+CoT       & \underline{75.6}         & \underline{48.9}          & 64.1          & 35.7            & 40.1           \\ \hline
{LLMs+KGs}    & \mymodel\ & \textbf{86.4}        & \textbf{62.9}          & \textbf{97.5}          & {99.1}            & \underline{96.3}          \\ \hline
\end{tabular}
\end{table*}


\subsection{Experiment Settings}

\textbf{Datasets.} In our experiments, we use three domain-specific datasets: MetaQA~\cite{zhang2018variational}, WebQuestionsSP (WebQSP)~\cite{yih2016value}, and Complex WebQuestions (CWQ)~\cite{talmor2018web}. MetaQA is a movie question-answering dataset consisting of 1-hop, 2-hop, and 3-hop questions. WebQuestionsSP features questions requiring up to 2 hops, while Complex WebQuestions includes questions that extend up to 4 hops. MetaQA is based on a movie knowledge graph, whereas both WebQSP and CWQ rely on Freebase as the underlying knowledge graph. Due to the large size of the original Freebase, we utilize a subset that aligns with those used in prior research.

\textbf{Baselines.} We evaluate \mymodel\ against different baseline methods categorized into 3 groups: 1) Embedding-based methods (GrafNet ~\cite{singh2021grafnet}, PullNet ~\cite{sun2019pullnet}), 2) Retrieval based methods (KV-Mem ~\cite{miller2016key}, EmbedKGQA ~\cite{sardana2020embedkgqa}, TransferNet ~\cite{shi2021transfernet}, BiNet ~\cite{liu2022joint}), 3) Large Language Models (ChatGPT ~\cite{achiam2023gpt}, CoT ~\cite{wei2022chain}).

\textbf{Implementations.} In our experiments, we use LLaMA2-Chat-7B as the backbone for the LLM. The results for all baseline methods are directly obtained from the original papers. We evaluate model performance using different key metrics: Precision (P), Recall (R), Hits@1 and F1. Hits@1 focuses on how frequently the top-ranked prediction is correct, offering a straightforward accuracy measure. On the other hand, F1 captures a broader picture by considering all possible correct answers, balancing precision and recall to reflect both accuracy and the completeness of the predictions.

\subsection{Main Results}

Table ~\ref{tableRes} presents the Hits@1 performance comparison across different question-answering (QA) datasets and methods. As we can see, \mymodel, outperforms other approaches, particularly on the WebQSP and CWQ datasets, achieving Hits@1 scores of 86.4\% and 62.9\%, respectively. Compared to the baseline methods, including traditional retrieval-based approaches like GraftNet and PullNet, and embedding-based methods such as KV-Mem and EmbedKGQA, \mymodel\ demonstrates superior performance.

While large language models (LLMs) such as ChatGPT and ChatGPT+CoT offer competitive results on some datasets, they fall behind on multi-hop reasoning tasks such as MetaQA, where retrieval-based methods and LLM+KG combinations excel. Our model (\mymodel), which integrates LLMs with knowledge graphs (KGs), consistently achieves competitive results on all datasets, demonstrating its ability to effectively combine both KG structure and language models' reasoning capabilities.

At the same time, \mymodel\ performs well on complex multi-hop tasks such as MetaQA-3hop, where it achieves a Hits@1 score of 96.3\%, highlighting its strength in handling longer reasoning chains.
We observed that all baselines perform very well on the MetaQA dataset, as the questions are straightforward and easy to answer.

\subsection{Ablation Study}

\begin{table}[hbt!]
\centering
\caption{Compare with RoG.}
\small
\label{table:rog}
\begin{tabular}{|l|c|c|c|c|c|}
\hline
\textbf{Dataset} & \multicolumn{5}{c|}{{WebQSP}}   \\ \hline
 & {P} & {R} & {F1} & H@1 & Acc \\ \hline
\mymodel\       & {69.20}              & {76.52}           & {67.29}  & \textbf{86.81} & \textbf{76.96}    \\
RoG  & \textbf{74.24}        & \textbf{76.71}        & \textbf{70.48}  & 86.29 & 76.48    \\ \hline
\end{tabular}
\end{table}
Table \ref{table:rog} shows that \mymodel\ outperforms RoG ~\cite{luo2024rog} in Hits@1 and accuracy, while RoG achieves better precision, recall, and F1 scores on the WebQSP dataset.
Table~\ref{table:performance} shows the results of an ablation study on two datasets, WebQSP and CWQ, comparing different versions of \mymodel\ based on precision, recall, and F1 score. The full version of \mymodel\ consistently delivers the best performance across both datasets, achieving the highest precision (69.20\% on WebQSP and 52.54\% on CWQ) and F1 scores (67.29\% on WebQSP and 54.85\% on CWQ). This highlights the effectiveness of incorporating both rule inference and re-ranking mechanisms, which substantially enhance the model’s prediction accuracy.

\begin{table}[hbt!]
\centering
\caption{Performance comparison of different versions of \mymodel\ on WebQSP and CWQ.}
\label{table:performance}
\small
\setlength{\tabcolsep}{3pt}
\begin{tabular}{|l|ccc|ccc|}
\hline
{\textbf{Model Variant}} 
& \multicolumn{3}{c|}{\textbf{WebQSP}} 
& \multicolumn{3}{c|}{\textbf{CWQ}} \\ \cline{2-7}
& P & R & F1 & P & R & F1 \\ \hline
\mymodel\           
& \textbf{69.20} & \underline{76.52} & \textbf{67.29}
& \textbf{52.54} & \underline{57.38} & \textbf{54.85} \\

\mymodel\ (no rule infer) 
& \underline{52.85} & 51.56 & 52.19
& 30.05 & 33.93 & 31.87 \\

\mymodel\ (no re-rank) 
& 41.48 & \textbf{80.45} & \underline{54.74}
& 14.59 & \textbf{66.94} & 23.96 \\

\mymodel\ (random rule) 
& 33.21 & 39.73 & 36.18
& \underline{33.17} & 38.35 & \underline{35.57} \\ \hline
\end{tabular}
\end{table}
Disabling rule inference leads to a noticeable drop in both precision and F1 scores, with precision falling to 52.85\% on WebQSP and 30.05\% on CWQ. This emphasizes the crucial role of rule inference in improving prediction precision, as its absence results in less accurate predictions.
When the re-ranking mechanism is removed, there is a trade-off between precision and recall. Although recall increases significantly (80.45\% on WebQSP and 66.94\% on CWQ), this comes at the cost of precision, which drops to 41.48\% and 14.59\%, respectively. The decline in F1 score reflects how removing re-ranking improves recall by retrieving more answers but decreases precision, leading to more false positives.
Using random rule selection results in the weakest performance, with F1 scores dropping to 36.18\% on WebQSP and 35.57\% on CWQ. This underscores the importance of carefully chosen rules, as random selection proves far less effective in guiding the model to accurate predictions.

\section{Related Work}

\textbf{Knowledge Graph Reasoning.} 
Knowledge graph reasoning has been studied for a long time
~\cite{liu2019g,liu2021neural,liu2021kompare,liu2022joint,liu2022comparative,liu2022knowledge,liu2023knowledge,liu2016brps,liu2024logic,liu2024can,liu2024new,liu2024conversational,liu2025transnet,liu2025neural,liu2025few,liu2025monte,liu2025hyperkgr,liu2025mixrag,liu2024knowledge,liu2025unifying,liu2026neural,liuneural,liu2026accurate,liu2026accurate2,liu2026ambiguous,liu2026ambiguous2}.
This problem typically involve three key factors: the input query $\mathcal{Q}$,
the reasoning model $\mathcal{F}$ and 
the Knowledge Graph $\mathcal{G} $\cite{liu2024new}.
Traditional KG reasoning approaches have primarily emphasized the model $\mathcal{F}$ and the knowledge graph $\mathcal{G}$. Reasoning strategies are usually categorized into three main types: embedding-based methods, rule-based methods and path-ranking methods \cite{liu2024new}.
{Embedding-based methods} map entities and relations into low dimensional vector spaces, where a score function is learned to predict the validity of entity-relation pairs. Prominent examples include \textbf{TransE} \cite{bordes2013translating} and \textbf{ComplEx} \cite{trouillon2016complex}. However, such methods struggle with some relational properties like symmetry, anti-symmetry, inversion and composition. While \textbf{RotatE} \cite{sun2019rotate} addresses these challenges by modeling relations as rotations in complex space, it still faces difficulties in generalizing across new datasets.

\textbf{Knowledge Graph Reasoning with Large Language Models.} Recent advancements in Large Language Models(LLMs) have prompted researchers come to explore whether these models could address some of the limitations of traditional KG reasoning methods. \textbf{GenTKG} \cite{liao2023gentkg} investigates LLMs(LLaMA2-7B for their experiment) as new foundational model of KG reasoning. Their result shows a significant reduction in the amount of training data required and a noticeable improvement in cross-domain generalizability, even when applied to Temporal Knowledge Graphs. However, their approach is limited by its reliance on single-shot logical rule inputs. \textbf{KG-GPT} \cite{kim2023kg} combines LLMs with graph retrieval, allowing the LLM to mark candidate relations. To address the lack of structural information in LLM-only approaches, \textbf{CSProm-KG} \cite{chen2023dipping} trains a graph encoder that integrates both textual and structural features before applying soft prompt tuning. However, as the model size increases, retraining and fine-tuning becomes very time consuming \cite{liu2024new}. This challenge has motivated the introduction of \textbf{Retrieval Augmented Generation(RAG)}, which leverages external knowledge sources. \textbf{ECOLA} \cite{han2022ecola} optimizes the knowledge-text prediction and knowledge embedding objectives simultaneously. Their newly constructed datasets contain contextual description for each temporal triplet, and future work may explore generating these contexts using LLMs. \textbf{REPLUG} \cite{shi2023replug} employs a retriever to fetch query related documents from the external corpora, concatenating them with context respectively. The retriever is optimized by matching the LLM ranking-based likelihood with the retriever's selection likelihood. This method helps keep the LLM frozen, reducing the need for costly retraining.



\section{Conclusion}

In this paper, we present a novel approach that integrates large language models (LLMs) with knowledge graphs to tackle the challenges of multi-hop question answering and hallucinations in LLMs. By leveraging logic rules and retrieval-augmented generation in the embedding space, our method demonstrates strong performance. Comprehensive empirical evaluations on multiple benchmark datasets confirm the effectiveness of our proposed method, \mymodel, in improving the performance of existing approaches.





\nocite{Ando2005,borschinger-johnson-2011-particle,andrew2007scalable,rasooli-tetrault-2015,goodman-etal-2016-noise,harper-2014-learning}

\bibliography{custom,liu}
\bibliographystyle{splncs04}

\appendix

\end{document}